\pdfoutput=1

\documentclass[11pt]{article}

\usepackage[]{EMNLP2023}

\usepackage{times}
\usepackage{latexsym}
\usepackage{graphicx}
\usepackage{enumitem}
\usepackage{hyperref}

\usepackage[T1]{fontenc}

\usepackage[utf8]{inputenc}

\usepackage{microtype}

\usepackage{inconsolata}
\usepackage{subcaption}
\usepackage{lipsum} 

\newcommand{\subsubsubsection}[1]{\noindent\textbf{#1}}

\title{MedLM: Exploring Language Models for Medical Question Answering Systems}

\author{%
  Niraj Yagnik\\
  UC San Diego \\
  nyagnik@ucsd.edu\\
    \And
   Jay Jhaveri \\
   UC San Diego \\
   jjhaveri@ucsd.edu \\
  \And
  Vivek Sharma \\
  UC San Diego \\
  v7sharma@ucsd.edu\\
  \And
  Gabriel Pila \\
  UC San Diego \\
  ypilahuancachoque\\
  @ucsd.edu \\
}

\begin{document}
 
\maketitle

\begin{abstract}
In the face of rapidly expanding online medical literature, automated systems for aggregating and summarizing information are becoming increasingly crucial for healthcare professionals and patients. Large Language Models (LLMs), with their advanced generative capabilities, have shown promise in various NLP tasks, and their potential in the healthcare domain, particularly for Closed-Book Generative QnA, is significant. However, the performance of these models in domain-specific tasks such as medical Q\&A remains largely unexplored. This study aims to fill this gap by comparing the performance of general and medical-specific distilled LMs for medical Q\&A. We aim to evaluate the effectiveness of fine-tuning domain-specific LMs and compare the performance of different families of Language Models. The study will address critical questions about these models' reliability, comparative performance, and effectiveness in the context of medical Q\&A. The findings will provide valuable insights into the suitability of different LMs for specific applications in the medical domain.
\end{abstract}

\section{Introduction}

With the plethora of online information on medical literature and research topics, it becomes incredibly challenging to keep up with the latest findings . The need for ubiquitous automated systems for aggregating and summarizing essential information would make it easier for healthcare professionals to apply evidence-based knowledge to their decision-making processes productively.
Similarly, from the patient's perspective, there is an increasing demand for easily-accessible accurate, and reliable medical information for the general population. An automated system would help patients with the information they can trust based on established journals and research to equip them to make educated and informed decisions for their health.

The latest advancements in generative models have made Large Language Models ubiquitous on the internet \cite{zhao2023survey}. The models are becoming increasingly large and complex, allowing them to learn intricate patterns available on the vast text available online to improve performance on a wide range of tasks. LLMs have achieved state-of-the-art results on various NLP tasks, including sentiment analysis, text summarization, and text generation. The capabilities of LLMs have been proven to extend beyond what was once considered challenging. This includes the creation of unique forms of text such as poems, code, scripts, musical compositions, emails, and letters \cite{sallam2023utility}. The last year has also seen an increased availability of these language models to the general public allowing professionals from all backgrounds to access them via their interface or API services for their specific use-case.

A lot of prior work on automated medical question answering is based on information retrieval systems, which work great for the task at hand but need to consider the specific context and nuances of the patients. There is a fundamental need for more personalization, which many modern automated medical systems need while trying to address patients. Large Language Models have presented themselves as reasons for considering the chain of thoughts and understanding the user context before answering the question. This makes it essential to leverage the strengths of generative language models and apply their power to the task of Closed-Book Generative QnA for the healthcare domain. 

The recent development in the NLP since the inception of transformers and attention mechanisms has seen the birth of multiple classes of language models to address and specialize different downstream tasks. There has been massive progress in the decoder-only family of models, with architectures like GPT-3 and GPT-4 \cite{radford2019language} producing outstanding results on a plethora of generative tasks. Similar improvement has been observed in the encoder-only class of families like the T5 \cite{raffel2020exploring} models have proved to be an efficient model for a diverse set of problems. This unprecedented makes it crucial to experiment with variations of language models from each class to deduce what works best for the task of generative question-answering

The plethora of medical information has led to the growing need for designing and developing automated systems that can assist laypeople and health workers find accurate answers to questions related to diagnosis, medications, treatment, side effects, etc. Medical Q\&A (Question Answering) systems have great potential to address this need but are still a massive work in progress. The efficacy of these systems largely relies on the scale and quality of the underlying language models used to generate answers. Similarly, large-scale pre-trained language models like GPT-3 and T5 have achieved outstanding performance in general language tasks; their performance on the domain-specific task of medical Q\&A is mainly unexplored. 

The proposed work aims to address the gap mentioned above in the literation by comparing the performance of general and medical-specific distilled LMs for the task of medical Q\&A. 
The work aims to evaluate whether fine-tuning domain-specific LMs leads to improved performance on the medical task compared to general LMs. The work also intends to provide a comparison across different families of  Language Models, i.e., a comparison of the performance of the decoder-only transformers model family(GPT-2, GPT-3) and the performance of the encoder-decoder models (BERT family) thus providing insights into the suitability of different LMs for this specific application.

The proposed work attempts to answer the following questions:
\begin{enumerate}
    \item Reliability Assessment of Generative Language Models for Medical QnA: Establishing Benchmark Scores for Closed Generative Question Answering in the Medical Domain.
    \item Comparative Performance Analysis of Distilled Fine-Tuned Language Models v/s General LLMs for Medical QnA.
    \item Performance Evaluation of Decoder-only Models versus Encoder-Decoder Models for Medical QnA.
    \item Effectiveness Evaluation of Prompt Engineering in Enhancing the Performance of LLMs for Medical Question and Answering Tasks.
\end{enumerate}

Making comparisons of the performance of general and medical-specific LMs on medical Q\&A tasks is necessary to determine which type is best suited for the application of medical domain. The comparisons would provide a deep insight into the strengths and limitations of different Language Models and help us highlight areas that need improvement. The comparisons would thus assist developers, researchers, and medical practitioners decide which model is the best for their specific task about Medical Q\&A.

\section{Literature Survey}

The task of Automated Question-Answering can be divided further into three variants based on the inputs and outputs \cite{huggingface}:
\begin{itemize}
    \item \textbf{Extractive QA:} The trained model extracts the answer from the provided context. The context can be text, table, image, or HTML. BERT-like models are generally the go-to approach to solving this problem. These models extract the span from the context which has the correct answer.

    \item \textbf{Open Generative QA:} Given the context, the model generates free text for the task. These models take the context as input and develop a coherent answer in natural language.

    \item \textbf{Closed Generative QA:} No context is provided for these models. It doesn't rely on external information and generates a response based solely on pre-trained knowledge. 
\end{itemize}

There has been some research on using Large Language models for medical question-answering tasks, detailed below:

BioBERT\cite{lee2020biobert} is a  pre-trained biomedical language representation model for biomedical text mining.
ClinicalBERT\cite{https://doi.org/10.48550/arxiv.1904.05342}talks about modeling Clinical Notes and Predicting Hospital Readmission". Kexin Huang, Jaan Altosaar, Rajesh Ranganath. This work develops and evaluates representations of clinical notes using bidirectional transformers (ClinicalBERT).

 Med-BERT\cite{Rasmy2021} is pretrained contextualized embeddings on large-scale structured electronic health records for disease prediction. This paper adapts the BERT framework originally developed for the text domain to the structured electronic health records EHR domain.
 
In "Can Large Language Models reason about medical questions?"\cite{https://doi.org/10.48550/arxiv.2207.08143} by Lievin et al. (2023), the paper evaluates the applicability of GPT-3.5 on medical questions by using different prompt strategies and later evaluated them with a medical expert. 

Med-PALM\cite{https://doi.org/10.48550/arxiv.2212.13138}: "Large Language Models Encode Clinical Knowledge"  by Karan Singhal, Shekoofeh Azizi et al. (2022). This paper proposes a framework to evaluate the large language  model answers along multiple axes including factuality, precision, possible harm, and bias.

 A medical question answering system using large language models and knowledge graphs\cite{Guo2022}.This study focuses on building a retrieval-based medical question answering system, tackling the challenge with large language models and knowledge extensions via graphs.

\section{Datasets}

To effectively assess various models, we conducted an evaluation based on datasets consisting of questions and answers used by the general public. Our selection process adhered to three specific criteria:

\begin{itemize}
\item  \textbf{Dataset task}: Among the plethora of text-based medical datasets available, we specifically focused on those centered around question and answer interactions, with a focus on being used for text generation. This ensured that our evaluation concentrated on datasets with relevant answers.

\item \textbf{Domain of the questions}: While certain datasets were limited to specific domains like COVID-related queries, we aimed to identify datasets that encompassed general questions that any patient might ask. This allowed for a comprehensive evaluation that addressed a broad range of medical inquiries.

\item \textbf{Trustworthiness}: We prioritized datasets that provided the most reliable and credible information. Accordingly, we favored datasets sourced from reputable institutions like the National Institutes of Health (NIH), ensuring the inclusion of trustworthy data in our evaluation.

\end{itemize}

Following a thorough assessment based on these criteria, we ultimately selected two datasets, namely MedQuAD \cite{BenAbacha-BMC-2019} and Icliniq \cite{LasseRegin/medical-question-answer-data}, to proceed with our analysis.

\begin{figure}[ht]
    \centering
    \includegraphics[width=1\linewidth]{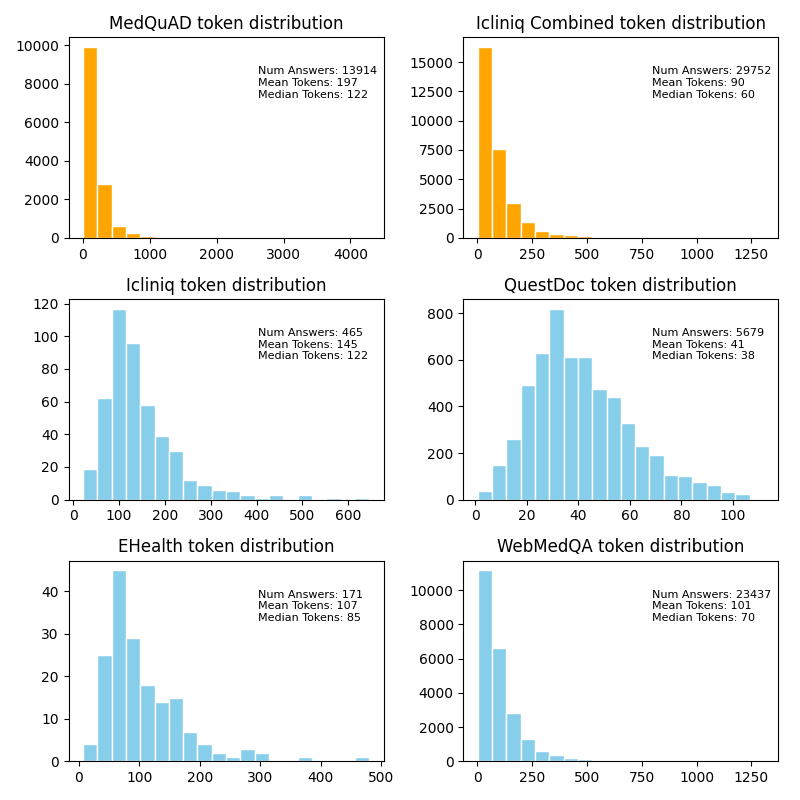
    }
    \caption{Distribution of token length in the datasets\\Note: We consider Icliniq dataset to the combination of the 4 online sources in blue}
    \label{fig:userReviewHeatmap}
\end{figure}

\subsection{MedQuAD}

MedQuAD is a comprehensive dataset comprising 47,457 question-answer pairs sourced from various NIH websites. The dataset stands out for its high-quality content, as the answers are supported by the authoritative backing of the NIH. MedQuAD encompasses an extensive range of medical inquiries, spanning 37 distinct question types. These questions encompass diverse areas such as treatment, diagnosis, and other pertinent aspects related to diseases, drugs, and various medical entities.
This dataset's answers had a length that varied from 50 words to more than 4,000 words, however, the most relevant information usually appeared in the first part of the answer, and the rest of the response was complimentary.

\subsection{Icliniq}

The second dataset is a compilation of 29,752 question-answer pairs collected from prominent websites such as eHealth Forum, iCliniq, Question Doctors, and WebMD. These sources contribute to the dataset's diversity and coverage of medical information. By including data from multiple reputable platforms, the dataset offers a broader perspective on various medical topics and enhances the overall comprehensiveness of the collection.

\section{Methodology}
Our research methodology encompassed three key procedures to analyze and improve the performance of large language models (LLMs). The methods employed included testing base LLMs, finetuning distilled versions of LLMs, and employing in-context learning via prompting of base LLMs.

\subsection{Finetuning Distilled Versions of LLMs}

\begin{figure}[ht]
    \centering
    \includegraphics[width=1.1\linewidth]{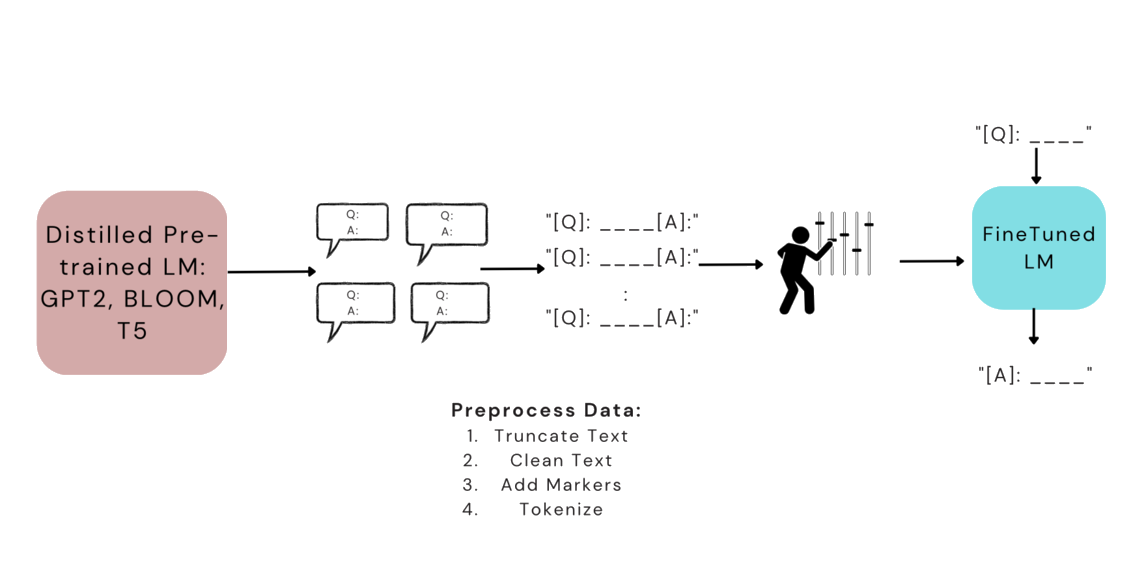}
    \caption{Finetuning Distil Models}
    \label{fig:fineTuningProcess}
\end{figure}

The first phase of our methodology involved fine-tuning the distilled models. In Fine-tuning, pretrained LLMs are further trained on specific tasks, here QnA, to allow the model to adapt its previously learned knowledge to the new task. This procedure involved multiple steps shown in Fig. \ref{fig:fineTuningProcess}.

\begin{flushleft}
\begin{itemize}[label=\textbf{-}]
    \item \textbf{Concatenate questions and answers:} Since our objective was to develop a generative model, we concatenated the question and answer pairs into a single text sequence. This allowed the model to learn the relationship between questions and answers, enabling it to generate appropriate responses when presented with a question.

    \item \textbf{Truncate model input:} Due to the varying lengths of the answers in the datasets, we standardized the input length by truncating it to a maximum of 300 tokens for the MedQuAD dataset and 150 tokens for the Icliniq dataset. Our decision to select the first 300 tokens was based on the observation that the most relevant content of each answer was typically found at the beginning. Considering that the mean answer length was 197 words, choosing 300 tokens allowed us to include the necessary tokens for the question as well. Additionally, by truncating the answer length, we reduced the processing time required to train the models, as longer inputs required more time for model computations.

    \item \textbf{Tokenize and fine-tune:} Once the input was truncated, we proceeded with the specific tokenization steps required for each model architecture. After tokenization, the models were fine-tuned using the information from the collected datasets. By following this process, we aimed to optimize the generative models' performance by fine-tuning them on the concatenated question-answer pairs, standardizing input length, and utilizing appropriate tokenization techniques for each model architecture.

\end{itemize}
\end{flushleft}

Following this, performance testing was conducted on the test set. As part of our methodology, we also employed a data augmentation technique where models were trained on both the MedQuad and iClinic datasets, thus broadening the diversity of the training data and potentially improving model robustness.

To accommodate the unique response types of each dataset, we adjusted the training inputs accordingly. Specifically, inputs were cut off at 300 words for the MedQuad dataset and 150 words for the iClinic dataset.

\subsection{Testing Base LLMs}

 Following the fine-tuning method, we performed a comprehensive testing of base LLMs, which involved the pretrained models provided by OpenAI from the decoder-only family, GPT2 GPT3.5 and Bloom, and by Google from the Encoder-decoder family, T5-base. These LLMs were tested on a diverse range of question-answering inputs present in the MedQuad and iClinic datasets. The models were tested to cover various aspects of answer generation capabilities, ensuring a broad coverage of topics and complexities.

\subsection{Prompting Base LLMs (In-Context Learning)}
To further explore the capabilities of the base LLMs, we implemented in-context learning, a technique that uses prompting to guide the models' responses. The base models were provided with prompts that included a set of examples indicative of the desired behavior. These examples provided the models with contextual information to guide their generation. We explored a variety of prompt designs, which included static prompting and dynamic prompting as explained below.

\subsubsection{Static Prompting}
In static prompting, a fixed set of two question-answer pairs were selected randomly from the training set and used as prompts before querying the model. This approach served to give the model a basic context for its response generation.

As we will see in the result section, this prompt worked great for some specific questions but performed far worse in other questions. Hence, we needed a process that will change the given prompt questions dynamically depending on the input test question.

\subsubsection{Dynamic Prompting}

Dynamic prompting, on the other hand, is a more refined technique. Here, the question-answer pairs used as prompts were selected based on their relevance to the query.
\\

\begin{figure}[ht]
    \centering
    \includegraphics[width=1\linewidth]{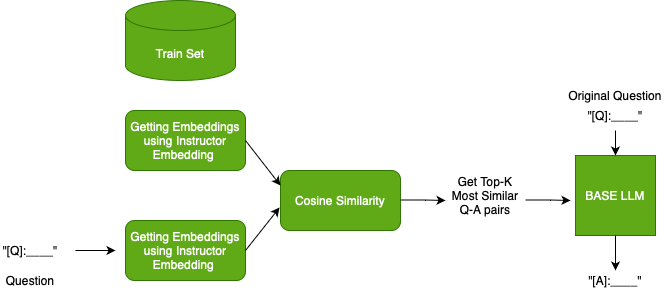}
    \caption{Dynamic Prompting}
    \label{fig:dynamicPromptingProcess}
\end{figure}

\subsubsubsection{Vanilla Dynamic Prompting\\}
The first approach we explored is called Vanilla Dynamic Prompting. Our plan was to embed all the questions available in the training set of the Medquad dataset and then calculate the cosine similarity between the embedded vectors and the input test question, ensuring that the selected prompts are highly related to the query, potentially improving the model's ability to generate a suitable response. This process \ref{fig:dynamicPromptingProcess} helped us identify the top $k$ questions to be used as prompts.

To embed these questions, we utilized the InstructOR Embedder \cite{su2022one} model. Unlike classic embedders, InstructOR Embedder accepts a domain sentence as input, which makes the embedded vectors more suitable for the specific domain and task at hand. In our case, we used the domain sentence: "Represent the Medicine sentence for retrieval: ".

During the inference phase, we employed the input test question to identify the top $k$ training questions. Through ablation studies, we determined that setting $k = 2$ yielded the best results. We then utilized these two questions, along with their corresponding answers, as prompts for performing in-context learning with large language models (LLMs). Finally, we incorporated the actual test question as the third question in the prompt.

Although this approach showed some improvement in scores, the gains were not significant. We concluded that the vastness of available training questions played a role in limiting its effectiveness. Therefore, we needed a way to categorize or filter the training questions.
\\

\begin{figure}[ht]
    \centering
    \includegraphics[width=1\linewidth]{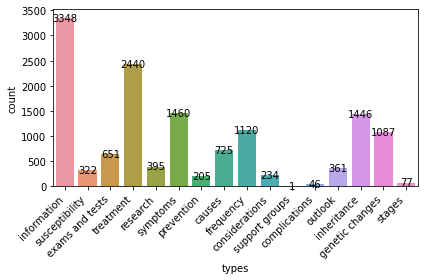}
    \caption{Various Question-Types}
    \label{fig:qtypes}
\end{figure}

\subsubsubsection{Question-Type Specific Dynamic Prompting\\}
In order to address the limitation of Vanilla Dynamic Prompting, we developed a more sophisticated technique called Question-Type Specific Dynamic Prompting. We leveraged the knowledge that the Medquad training set included question types associated with each question, such as "Symptoms," "Treatments," and "Information," among others. In total, there were 16 different question types \ref{fig:qtypes}. Note: We did notice the imbalance in question types, but we are going to leave handling that for future work.

As a preprocessing step before inference, we trained a BERT classifier on the training data, where the input was the question, and the prediction variable was the corresponding question type. This classifier allowed us to determine the question type for a given input question.

Similar to the previous approach, we employed the InstructOR Embedder to create 16 separate embedder blocks, each representing one of the question types.

During the inference phase, we first used the pretrained BERT classifier to infer the question type of the incoming question. Next, we accessed the specific embedding block associated with the inferred question type and calculated the cosine similarity to obtain the top $k$ similar questions. Again, we found that setting $k = 2$ produced the best results.

As we will discuss in the Results section, this Question-Type Specific Dynamic Prompting approach yielded a significant performance boost compared to the Vanilla Dynamic Prompting approach.

\section{Results}

The results of our study are presented in this section, where we evaluate the performance of the model both quantitatively and qualitatively. For automated evaluation, we employed the BLEU \cite{papineni2002bleu} and ROUGE \cite{lin2004rouge} metrics. The BLEU score measures the structural accuracy of generated sentences, while the ROUGE score assesses the extent to which the generated answers capture the overall meaning conveyed in the reference text. However, we found that these quantitative metrics alone were not entirely reliable, necessitating the need for human evaluation. To this end, we conducted surveys with health professionals and potential patients/users of MedLM, as our research focuses on a Medical QnA dataset.

We begin by examining the worst-performing questions. One such question is, "What is fever?" Our analysis reveals that each model possesses its own characteristics. The Bloom model, although factually correct, tends to repeat itself. GPT2 occasionally exhibits hallucinatory behavior by generating seemingly plausible yet fictional paragraphs. Moreover, the T5 model sometimes provides incorrect information. For instance, for this particular question, T5 falsely claimed that fever is a rare condition.

Next, when we explored the performance of large base language models, we found that these models exhibit sensitivity to prompts, particularly concerning hallucinations and answer length.

\subsection{Quantitative Results}

Moving on to the results obtained from testing the models on the Medquad dataset's test set \ref{tab:Results:MedquadTest}, we assign the highest importance to the ROUGE-1 metric as it signifies the effectiveness of conveying the expected information accurately. Our analysis indicates that GPT 3.5 with static prompts yields the best results. Notably, the fine-tuned models do not surpass these scores but achieve comparable performance.

\begin{table*}[t]
\centering
\small 
\setlength{\tabcolsep}{4pt} 
\begin{tabular}{@{}lcccc@{}}
\hline
Model & Bleu1 & Bleu4 & Rouge-1 & Rouge-L \\ \hline
T5 & 6.186 & 0.239 & 0.199 & 0.178 \\
GPT-2 & 3.806 & 0.132 & 0.196 & 0.181 \\
Bloom & 1.577 & 0.043 & 0.193 & 0.18 \\
GPT-3.5 w/Static Prompt & 3.413 & 0.122 & \textbf{0.232} & 0.216 \\
GPT-3.5 w/TypeWise Dynamic Prompt & 7.132 & 0.344 & 0.222 & 0.211 \\
ChatGPT-4 Dynamic Prompting & 3.55 & 0.107 & 0.129 & 0.121 \\ 
GPT-3.5 w/TypeWise Dynamic Prompt &	7.132 &	0.344 &	0.222 &	0.211 \\
ChatGPT-3 Dynamic Prompting	& 5.5287 & 0.290 &	0.2029 & 0.1897 \\ \hline
\end{tabular}
\caption{Model Evaluation On Medquad Testset}
\label{tab:Results:MedquadTest}
\end{table*}

In accordance with the above observation, we investigate the impact of data augmentation. To explore this, we further trained the finetuned distilled models on the icliniq dataset and evaluated them on the medquad dataset. Excitingly, the Bloom model performs almost as well as GPT 3.5 did in the previous analysis Table \ref{tab:Results:DataAug}. This finding suggests that incorporating more data and higher quality data improves the fine-tuning process.

\begin{table}[ht]
\centering
\begin{tabular}{lcccc}
\hline
Model & Bleu1 & Bleu4 & Rouge-1 & Rouge-L \\ \hline
T5 & 7.117 & 0.321 & 0.207 & 0.186 \\
GPT-2 & 5.132 & 0.242 & 0.190 & 0.174 \\
Bloom & 1.871 & 0.046 & 0.226 & 0.212 \\ \hline
\end{tabular}
\caption{Model Evaluation after Data Augmentation}
\label{tab:Results:DataAug}
\end{table}

In a separate experiment, we exclusively fine-tuned the distilled models on the icliniq dataset and evaluated their performance Table \ref{tab:Results:icliniqTest}. As expected, without the strong foundation provided by the medquad dataset, the icliniq performance is significantly poorer compared to GPT 3.5.

\begin{table*}[t]
\centering
\small 
\setlength{\tabcolsep}{4pt} 
\begin{tabular}{@{}lcccc@{}}
\hline
Model & Bleu1 & Bleu4 & Rouge-1 & Rouge-L \\ \hline
GPT3.5 w/Static Prompt & 2.434 & 0.097 & 0.221 & 0.199 \\
T5 & 9.843 & 0.807 & 0.173 & 0.145 \\
GPT-2 & 9.989 & 0.855 & 0.139 & 0.123 \\ \hline
\end{tabular}
\caption{Model Evaluation On icliniq Testset}
\label{tab:Results:icliniqTest}
\end{table*}

\subsection{Qualitative Results}

Shifting our focus to qualitative results, we conducted a survey with potential patients/users to gauge the comprehensibility of the generated answers in natural language. The survey results, calculated using the Likert scale \cite{likert1932technique}, are seen in Fig. \ref{fig:userReviewHeatmap}. Surprisingly, some users preferred the answers generated by GPT models over the human-written ground truth answers. Additionally, the Bloom model consistently performs at the level of the ground truth, reaffirming our earlier observations.

\begin{figure}[ht]
    \centering
    \includegraphics[width=1.1\linewidth]{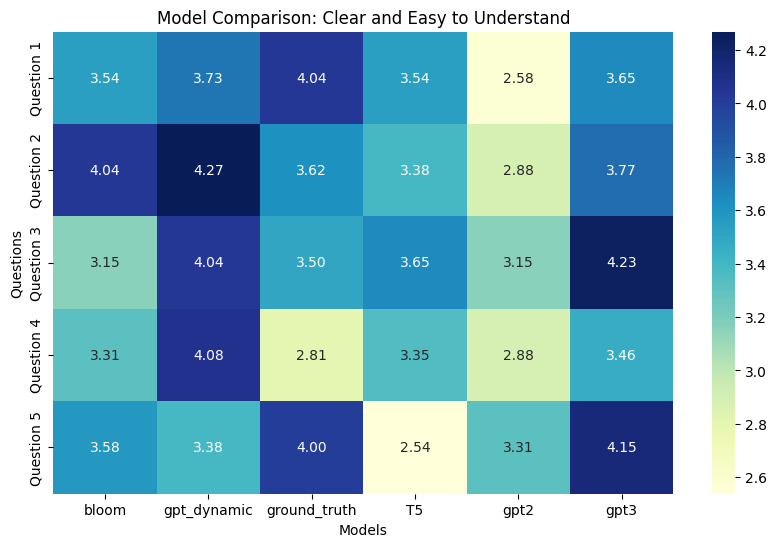}
    \caption{Heat Map of User Responses.\\Note: All survey scores calculated using Likert Scale \cite{likert1932technique}}
    \label{fig:userReviewHeatmap}
\end{figure}

\begin{figure}[ht]
    \centering
    
    \begin{subfigure}{1\linewidth}
        \centering
        \includegraphics[width=\linewidth]{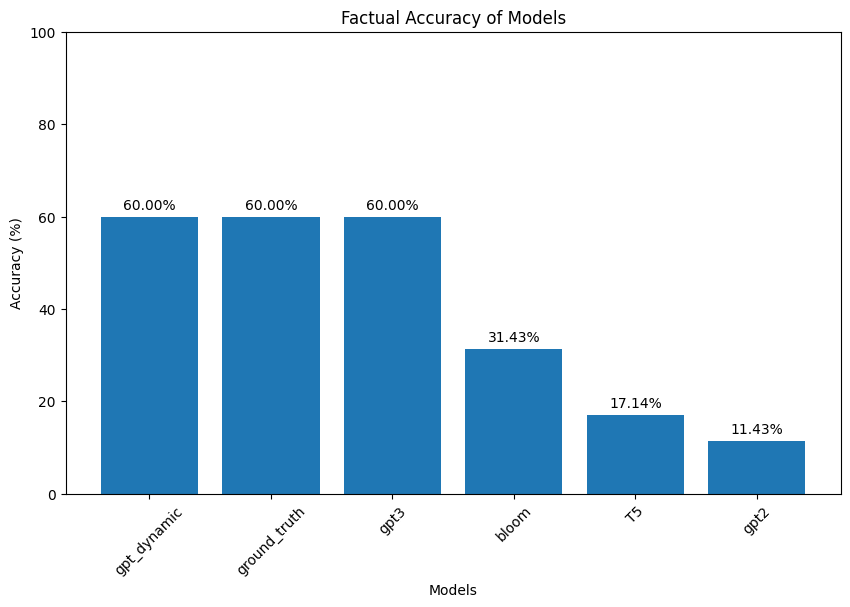}
        \caption{Factual Accuracy}
        \label{fig:FactualAcc}
    \end{subfigure}
    \hfill
    \begin{subfigure}{1\linewidth}
        \centering
        \includegraphics[width=\linewidth]{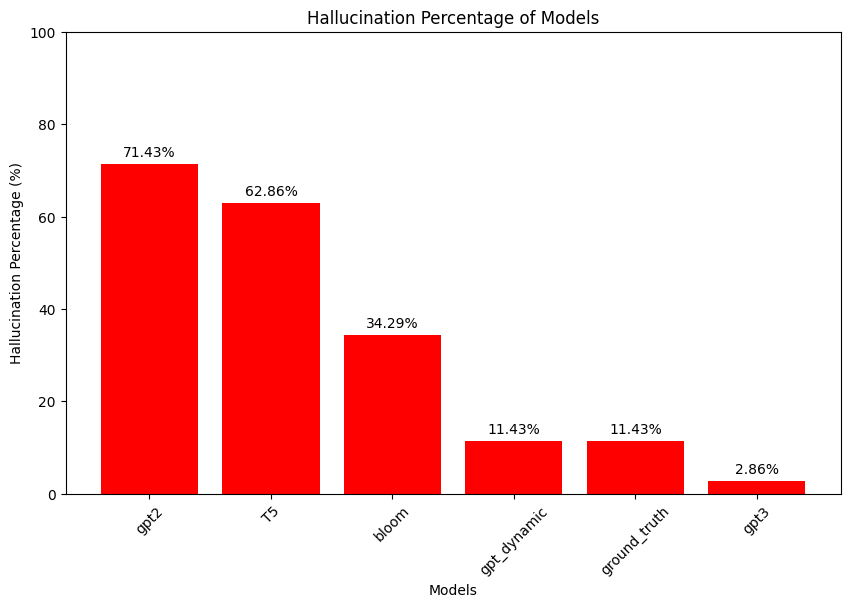}
        \caption{Hallucination}
        \label{fig:Halucination}
    \end{subfigure}
    
    \caption{Results of Doctor Survey using Liker Scale}
    \label{fig:DocSurvey}
\end{figure}

Moreover, we surveyed health professionals to evaluate the factual accuracy of the generated answers \ref{fig:DocSurvey}. The top graph \ref{fig:FactualAcc} in our analysis illustrates that, similar to the users' perspective, doctors also consider large GPT models to be consistently more factually reliable than the human-generated ground truth answers. The bottom graph \ref{fig:Halucination} indicates the propensity of models to hallucinate, with smaller fine-tuned models exhibiting higher rates of hallucination. Interestingly, doctors also believe that human-generated answers contain a certain degree of hallucination.

\subsection{Observation}

Throughout our research journey, we made several noteworthy discoveries. This study establishes a robust benchmark for future researchers working on medical QnA and other domain-specific tasks. Contrary to our initial expectations, the metrics employed for evaluation proved unreliable, and human evaluation yielded different conclusions. Decoder-only models, such as GPT, demonstrate superior performance. Additionally, our findings indicate that while static prompts yield better results for certain questions, they can lead to worse answers for others. Consequently, we introduced dynamic prompting, which consistently improved the results.

Overall, our research sheds light on the strengths and weaknesses of various models in the context of Medical QnA. The combination of quantitative and qualitative evaluations provides a comprehensive understanding of their performance and highlights avenues for future improvements.

\section{Limitations}
The work showcases that hallucination is prevalent in the answers generated, especially in the case of fine-tuned models of T5, BLOOM, and GPT-2. The hallucination could be resolved with the proper augmentation of data. Currently, the dataset has limited samples to fine-tune and often has varying answer lengths, with a few questions having incredibly verbose responses and others presenting only a few sentences.

Getting human evaluation for the responses generated by each model is challenging to scale. The current human evaluation of doctors and users was performed to evaluate the accuracy and understandability of 5 questions on the answers generated by six models/techniques the work experiments with. Getting human evaluations would be much more challenging to scale for higher examples and more models as a standard technique for evaluation. Participants who took the survey for the assessment expressed concerns about the length of the survey form.

Resource and computational constraints prevented us from further experiments on the fine-tuned models, like training with more epochs and larger batch sizes.

\section{Future Work}

The research paper proposes several future directions for further investigation and improvement:

\textbf{Test DynamicPromting on GPT-4 API:} For this project's scope, the work has only evaluated the dynamic prompting technique on the GPT3.5 API. By experimenting with different prompts and contextual cues on newer and better APIs, we aim to improve the accuracy and reliability of the generated responses.

\textbf{Fine-Tuning on GPT3, GPT-4, and other larger models:} Currently, the work fine-tunes the distilled versions of GPT-2, BLOOM, and T5 Models. The lack of availability of the refined open-source GPT-3 and GPT-4 models prevented the work from further experimentations. We propose exploring the impact of fine-tuning on these models to enhance their performance and reduce hallucination. Fine-tuning newer models may offer improved contextual understanding and generate more accurate answers.

\textbf{Coming up with better metrics for Generative QnA tasks:} As seen in the result section, models which produced accurate and understandable results according to human evaluations performed relatively poorly on the rogue and bleu metrics. Improved evaluation metrics will enable a more comprehensive and precise assessment of the models' performance and help identify and address hallucinations.

\textbf{Enhance the dataset through processing, augmentation, and summarization:}  As illustrated by the work, data augmentation is an effective technique and direction on which future work could focus. Another approach worth exploring is processing the answers to summarize them to make the dataset more uniform to have more consistent and cohesive results.

\section{Conclusion}

The proposed work lays an excellent benchmark for future work on Generative Closed-Book Question Answering. The work explores ten techniques across two datasets to determine what works best for the healthcare domain. Future work can use this as a stepping stone to explore more techniques mentioned beyond this work.

The work also exposes the sensitivity of prompting for Medical QnA. The static prompts technique does not perform as well compared to the method where the Base LLM generated the answers without prompting. The work highlights more advanced techniques for Dynamic prompting and how searching for questions of similar topics and contexts from the train set for prompting can drastically improve the metric scores and the output quality.

The work also highlights the efficacy of data augmentation as a technique to compensate for the relatively low volume of good-quality question-answer pairs. Last, the work highlights the need for new metrics beyond Rouge and Bleu to do justice to the answers given by generative models.

\section*{Code: Github}
The code for the proposed work can be found on our Github Repo:  \href{https://github.com/JayJhaveri1906/CSE291\_MedLM}{MedLM}

\section*{Ethics Statement}
Our research adheres to ethical guidelines and considerations. We have taken steps to ensure privacy and data protection by obtaining necessary permissions and informed consent. We have also made efforts to mitigate biases and evaluate the potential impact on different demographic groups. Transparency and interpretability are prioritized, providing clear explanations of our approach and limitations. We are committed to a safe and inclusive environment and consider the broader societal implications of our work. We welcome feedback and strive for continuous improvement in our ethical practices.

\section*{Acknowledgements}
We want to express our sincere gratitude to Dr. Jingbo Shang, Bill Hogan, and Dr. Asma Ben Abacha for their invaluable guidance and mentorship throughout this project. Their expertise and support have been instrumental in our progress. Without their contributions, our project wouldn't have reached its current stage. We are truly thankful for their valuable insights and unwavering support, which have greatly influenced our journey and outcomes.

\bibliography{main}
\bibliographystyle{acl_natbib}




\end{document}